\documentclass[]{interact}

\usepackage{epstopdf}
\usepackage{subfigure}

\usepackage{natbib}
\usepackage{amsmath,amsfonts}
\usepackage[ruled]{algorithm2e}
\usepackage{array}
\usepackage{booktabs}
\usepackage{multirow} 
\usepackage{url}
\usepackage{color}
\usepackage{hyperref}
\bibpunct[, ]{(}{)}{;}{a}{}{,}
\renewcommand\bibfont{\fontsize{10}{12}\selectfont}

\theoremstyle{plain}

\theoremstyle{definition}

\theoremstyle{remark}

\begin{document}

\articletype{ARTICLE}

\title{Evaluation of Drivers' Interaction Ability at Social Scenarios: A Process-Based Framework}

\author{\name{Jiaqi Liu\textsuperscript{a}, Peng Hang\textsuperscript{a}, Xiangwang Hu\textsuperscript{b}\thanks{CONTACT XiangwangHu. Email: xwhu@suda.edu.cn}  and Jian Sun\textsuperscript{a}
}
\affil{\textsuperscript{a}the Department of Traffic Engineering and Key Laboratory of Road and Traffic Engineering, Ministry of Education, Tongji University, Shanghai 201804, China; 
\textsuperscript{b}the School of Rail Transportation, Soochow University, No. 8 Jixue Road, Suzhou, Jiangsu, China}
}

\maketitle

\begin{abstract}
Assessing drivers' interaction capabilities is crucial for understanding human driving behavior and enhancing the interactive abilities of autonomous vehicles. In scenarios involving strong interaction, existing metrics focused on interaction outcomes struggle to capture the evolutionary process of drivers' interactive behaviors, making it challenging for autonomous vehicles to dynamically assess and respond to other agents during interactions. To address this issue, we propose a framework for assessing drivers' interaction capabilities, oriented towards the interactive process itself, which includes three components: Interaction Risk Perception, Interaction Process Modeling, and Interaction Ability Scoring. We quantify interaction risks through motion state estimation and risk field theory, followed by introducing a dynamic action assessment benchmark based on a game-theoretical rational agent model, and designing a capability scoring metric based on morphological similarity distance. By calculating real-time differences between a driver's actions and the assessment benchmark, the driver's interaction capabilities are scored dynamically. We validated our framework at unsignalized intersections as a typical scenario. Validation analysis on driver behavior datasets from China and the USA shows that our framework effectively distinguishes and evaluates conservative and aggressive driving states during interactions, demonstrating good adaptability and effectiveness in various regional settings.
\end{abstract}


\section{Introduction}
In future human-machine mixed driving environments, autonomous vehicles (AVs) must understand the social intentions of other drivers and navigate in a manner that aligns with human expectations\citep{liu2024enhancing,duo2023intra}. Comprehending and evaluating the interaction behaviors of drivers is essential for AVs to predict human drivers' future states and generate recognizable social driving behaviors \citep{parkin2016understanding,liu2023towards}. However, discerning human-driver interactions on the road is challenging for AVs, as humans do not strictly adhere to traffic rules like machines. Instead, they follow implicit social norms and rules to achieve safety and efficiency gains \citep{wang2022social,liu2023teaching}. Therefore, traditional metrics and rigid traffic rules cannot be used to evaluate their behaviors.

In recent years, Time to Collision (TTC) \citep{fiorentino1997time} and Post Encroachment Time (PET) \citep{qi2020modified} have been widely employed to assess and analyze vehicle interactions. Although these indicators can reflect consequential information, such as interaction event safety after completion, they do not assess the interaction process or the abilities of both parties \citep{markkula2020defining}. Data-driven methods like deep neural networks excel in quantifying interactions but suffer from limited interpretability and struggle to provide reliable evaluation criteria \citep{liu2023mtd,wang2022social,liu2024ddm}. Model-based methods, on the other hand, offer clearer structures and better interpretability. Game theory, in particular, is widely used to analyze interaction processes and model interactions \citep{li2022human,liu2024enhancing}. Game models can accurately capture drivers' dynamic interactions and provide a foundation for describing the spatiotemporal characteristics of driving behaviors.

To address these challenges,  we propose a process-oriented framework for evaluating drivers' abilities in dynamic and interactive scenarios, comprising three stages: \textbf{Risk Perception Modeling}, \textbf{Interaction Process Modeling}, and \textbf{Interaction Ability Scoring}. This framework is designed to assess social interaction behaviors in road traffic, particularly in complex interaction scenarios.

In our framework, we initially quantify interaction risks through motion state estimation combined with risk field theory. Subsequently, we introduce a dynamic action assessment benchmark using a game-theoretical rational agent model. This model employs both non-cooperative and cooperative games to simulate the action responses of drivers during interactions. By manipulating the weights in the game's payoff function, we can model rational agents with diverse behavioral styles.
After assessing a driver's perceived risk, the rational person model formulates benchmarks for various driving behaviors tailored to different social preferences. The evaluation of a driver's social competencies—such as safety, efficiency, and social interaction—is conducted by measuring the discrepancy between their actual behavior and these optimal benchmarks.

We have applied our framework to scenarios involving unprotected left turns in both China and the United States, allowing for a comparative analysis of driving abilities across different datasets. Validation analysis of driver behavior datasets from these regions confirms that our model effectively distinguishes between conservative and aggressive driving behaviors during interactions. This underscores its adaptability and effectiveness in diverse regional contexts.

The contributions of this paper are as follows:
\begin{itemize}
    \item We propose a three-stage ability evaluation framework for challenging and interactive scenes, including Risk Perception Modeling, Interaction Process Modeling, and Interaction Ability Scoring. This process-based framework is capable of analyzing drivers' performance and revealing what occurred during the interaction, compared to existing evaluation indicators.
    \item In the Interaction Process Modeling stage, we establish a rational person model based on a game theoretic approach. The rational person confronts interaction conditions similar to real-world drivers and generates decision-making results. By examining the differences between real-world drivers and the rational person, we can detect and evaluate driving characteristics during the interaction.
    \item In our framework, we develop a comprehensive risk perception model that combines instantaneous state risk and future state risk in traffic scenarios. The Extended Kalman Filter (EKF) is utilized to estimate the motion states of feature vehicles. Additionally, we propose a novel improved morphological similarity evaluation index to measure the gap between reality and rationality. Furthermore, we select the unprotected left-turn scenario as a typical testing scenario and assess the social abilities of Chinese and US drivers using different datasets.
\end{itemize}

The remainder of the paper is organized as follows: Section \ref{section:review2} summarizes recent related works. Section \ref{section:overview3} presents an overview of the proposed evaluation framework. Section \ref{section:design4}provides detailed information about the framework. Section \ref{section:exper5} describes the evaluation experiments using different datasets and analyzes the results. Finally, Section \ref{section:conclu6} concludes the paper.

\section{Related Works}
\label{section:review2}

\subsection{Driving Interaction Behavior Modeling and Assessment}
Constructing a model that accurately describes the interactive process of drivers is essential for evaluating interaction behavior. 
There are different kinds of methods to model the human driving behavior, such as reinforcement learning\citep{peng2021end,liu2023cooperative} with attention mechanism\citep{liu2023optimal,liu2024delay}, which is a powerful data-driven method, and game theory methods\citep{hang2020human,liu2024enhancing}. Compared with data-driven methods, game-based methods have the advantages in stability and interpretability, which have been widely used for behavior modeling in recent years\citep{wang2022social,liu2024enhancing}. 

In game theory, researchers have constructed various specific game models based on the assumption of absolute rationality to explain and assess the decision-making characteristics of drivers. Nash equilibrium is achieved when no player can unilaterally increase their expected payoff by changing their strategy. This can be divided into pure strategy games and mixed strategy games. In pure strategy games, drivers choose a particular strategy to obtain the optimal return\citep{ali2019game}. Mixed strategy games assign a probability to each driver's strategy\citep{arbis2019game}, with drivers selecting their strategies to maximize their own payoff based on the set probabilities.

However, it is important to note that in traditional environments, human drivers often struggle to accurately perceive their surroundings. Drivers tend to make decisions with bounded rationality, and the surrounding traffic environment along with other uncertain factors are often difficult to measure and capture accurately. Some scholars have acknowledged this factor and have explored it further\citep{chen2023game,arbis2019game}.

In game-based framework, human drivers are influenced by others to make decisions, and in turn, affect others, forming a closed-loop dynamic game. Models employing game theory to simulate driver-driver interactions have been widely used to evaluate and predict the driving behaviors of human drivers and design interaction strategies between autonomous vehicles and human-driven vehicles\citep{2020Game,2021Human,9536147,liu2024enhancing}. 
The performance of these models is typically assessed by examining the predicted trajectories of AVs.
However, human drivers' abilities are generally evaluated through driving simulation experiments and more detailed data analysis\citep{2000Characterological,2012MMSE}. Trajectory information alone is limited and cannot fully reflect drivers' intentions and actions during the interaction process or the driving abilities underlying these actions. Moreover, it is unrealistic for future large-scale, real-time interaction evaluation needs to be met by conducting driving simulation experiments. Therefore, exploring how to better utilize the output information of game models to evaluate drivers more effectively is a valuable endeavor.

\subsection{Risk Quantification and Evaluation}
Numerous methods have been employed to model the risk of vehicle motions. Existing risk assessment methods mainly encompass risk quantification indices, collision prediction, and risk field modeling.

Quantifying the risk of moving objects using indices based on time or distance is a common and effective approach. Time-based quantifying indices include Time to Collision (TTC), Time Headway (THW), Time Margin (TM), and Post Encroachment Time (PET)\citep{zhao2018risk,2019A,2019Potential,qi2020modified}, while distance-based indices commonly used are stopping distance and safety margin\citep{Doi1994Development,2017On,shalev2018vision}.

Evaluation methods based on collision prediction calculate the risk of collision by predicting feature trajectories of the interacting objects. The motion trajectories of vehicles are usually predicted using deep learning models or linear differential equations, and the intersection point between two trajectories can be efficiently computed\citep{2006A,2002An}.

In addition, field theory is a prevalent approach for assessing movement risks\citep{2020Dynamic}. The concept of Artificial Potential Field (APF), based on field theory, was first proposed by Khatib\citep{1986Real}. APF has also been used as a basis for calculating the total potential of a traffic environment\citep{2019Potential_2}.

Despite the widespread study and application of the methods mentioned above, conducting a more comprehensive risk assessment in complex interactive intersections remains a significant challenge.

\subsection{Performance Scoring and Similarity Evaluation}
The driver's interactive ability evaluation problem can be transformed into a motion similarity measurement problem between real-world drivers and excellent drivers. There are three dimensions for motion similarity measurement: 
(1) measuring spatial similarity, which focuses on geometric trajectory while ignoring the time dimension, such as Euclidean Distance\citep{little2001video,yuan2017review}; 
(2) measuring temporal similarity, which analyzes the temporal characteristics of sequence data, including cosine similarity\citep{Tetsuya2013A}, fractional order correlation\citep{li2022new}, and Symbolic-Aggregate Approximation (SAX)\citep{taktak2023novel} etc.; (3) measuring spatial-temporal similarity, which considers both spatial and temporal dimensions of motion data, with common methods like Dynamic Time Warping (DTW)\citep{deriso2023general,lahreche2021fast,2002Exact}.

However, single indices of spatial and temporal similarity evaluation mentioned above fail to capture many crucial features from drivers' action sequences. Additionally, spatial-temporal similarity evaluation indices, such as DTW, cannot normalize the differences between sequences into a finite interval, which is insufficient for meeting the requirements of action evaluation and scoring. Therefore, developing more comprehensive and effective evaluation methods that accurately capture the nuances of driver interactions while addressing the limitations of existing similarity indices remains a pressing challenge in the field.

\section{Framework Overview and Scenario Description}
\label{section:overview3}
In this section, the evaluation framework is outlined, and an unprotected left-turn scenario is introduced as a typical example to illustrate our framework in detail.

\subsection{Framework Overview}
\begin{figure}[!htbp]
    \centering
    \includegraphics[width=1 \textwidth]{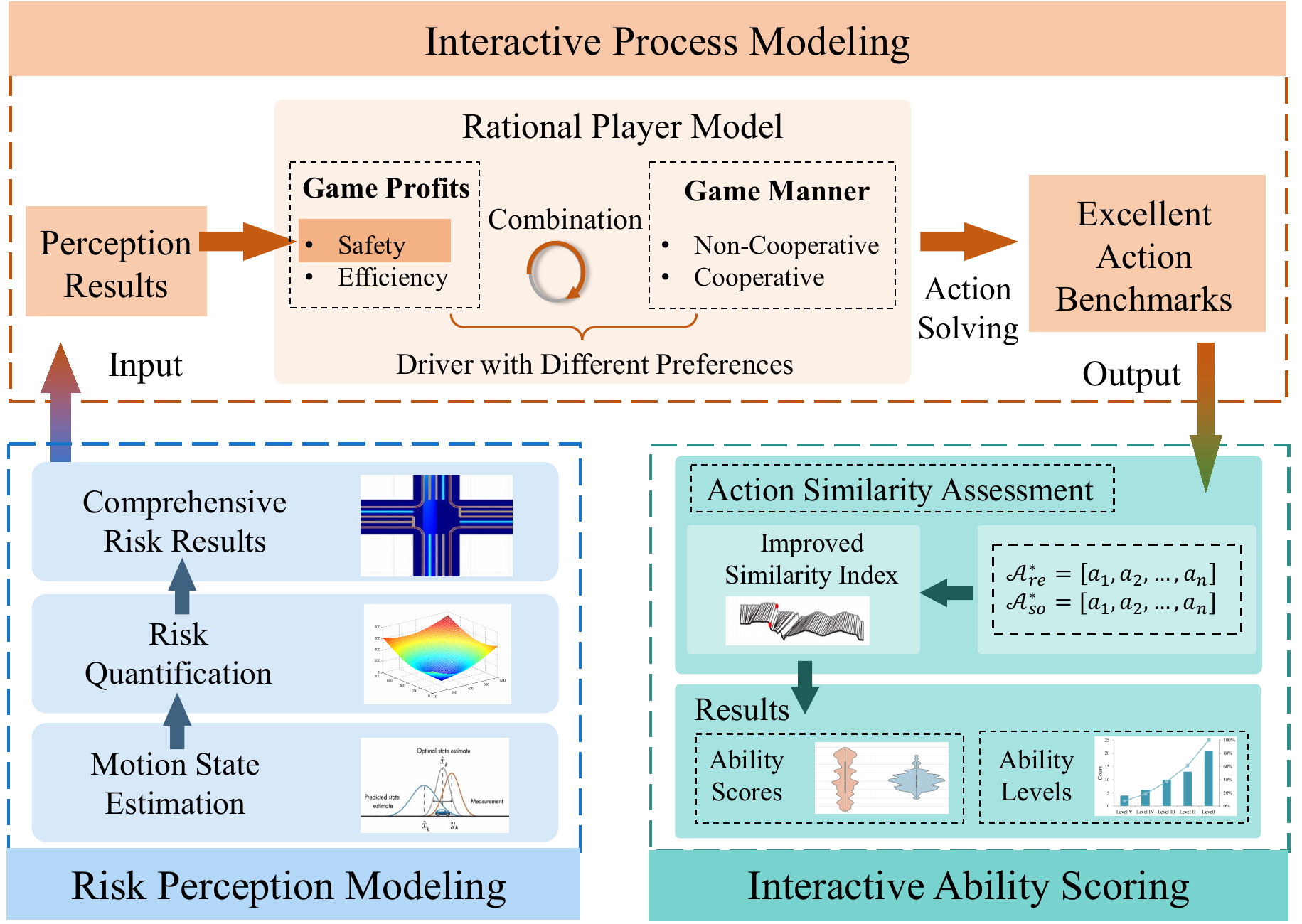}
    \caption{The framework of the interaction ability evaluation.}
    \label{fig:framework} 
\end{figure}

As illustrated in Figure \ref{fig:framework}, our proposed framework for assessing interaction capabilities consists of three components: Risk Perception Modeling, Interactive Process Modeling, and Interactive Ability Scoring. In the Risk Perception Modeling module, we employ an Extended Kalman Filter (EKF) for motion state estimation, model the risks associated with the current and future states using a hybrid approach, and quantify them based on risk field theory. Subsequently, in the Interactive Process Modeling section, we construct various types of rational decision-maker models based on game theory, using the decision outputs of these models as benchmarks for dynamic action assessment. In the Interactive Ability Scoring module, we have designed a scoring metric based on morphological similarity distance. In specific scenarios, this metric is used to evaluate the social interaction characteristics and decision-making abilities of drivers by real-time calculation of the differences between the decision outputs of the rational decision-maker models and the actual actions of real-world human drivers.
A detailed description of the three functional modules is presented as follows.
\begin{itemize}
    \item \textbf{Risk Perception Modeling}: By using motion state estimation with instantaneous states and EKF, the future states of drivers are obtained. The risk between drivers is calculated using risk field theory. The proposed risk model combines instantaneous state risk and future state risk, quantifying the risk of static and dynamic objects in the traffic environment.
    \item \textbf{Interactive Process Modeling}: The rational person model is first constructed, taking the results of risk perception as game safety benefits. The sum of safety benefits and efficiency benefits is used to construct the player's total game benefit function, and the rational player will make decisions and take actions based on the maximum game payoff at each time step. By adjusting the game model, rational players with different preferences are established.
    \item \textbf{Interactive Ability Scoring}: An improved morphological similarity evaluation index is proposed. The scores and rankings of drivers who participated in the experiments are obtained by measuring the action gaps between actions from real-world drivers and rational actions from game models with different preferences.
\end{itemize}

\subsection{Scenario Description}

\begin{figure}[htb]
    \centering
    \includegraphics[width=0.8 \textwidth]{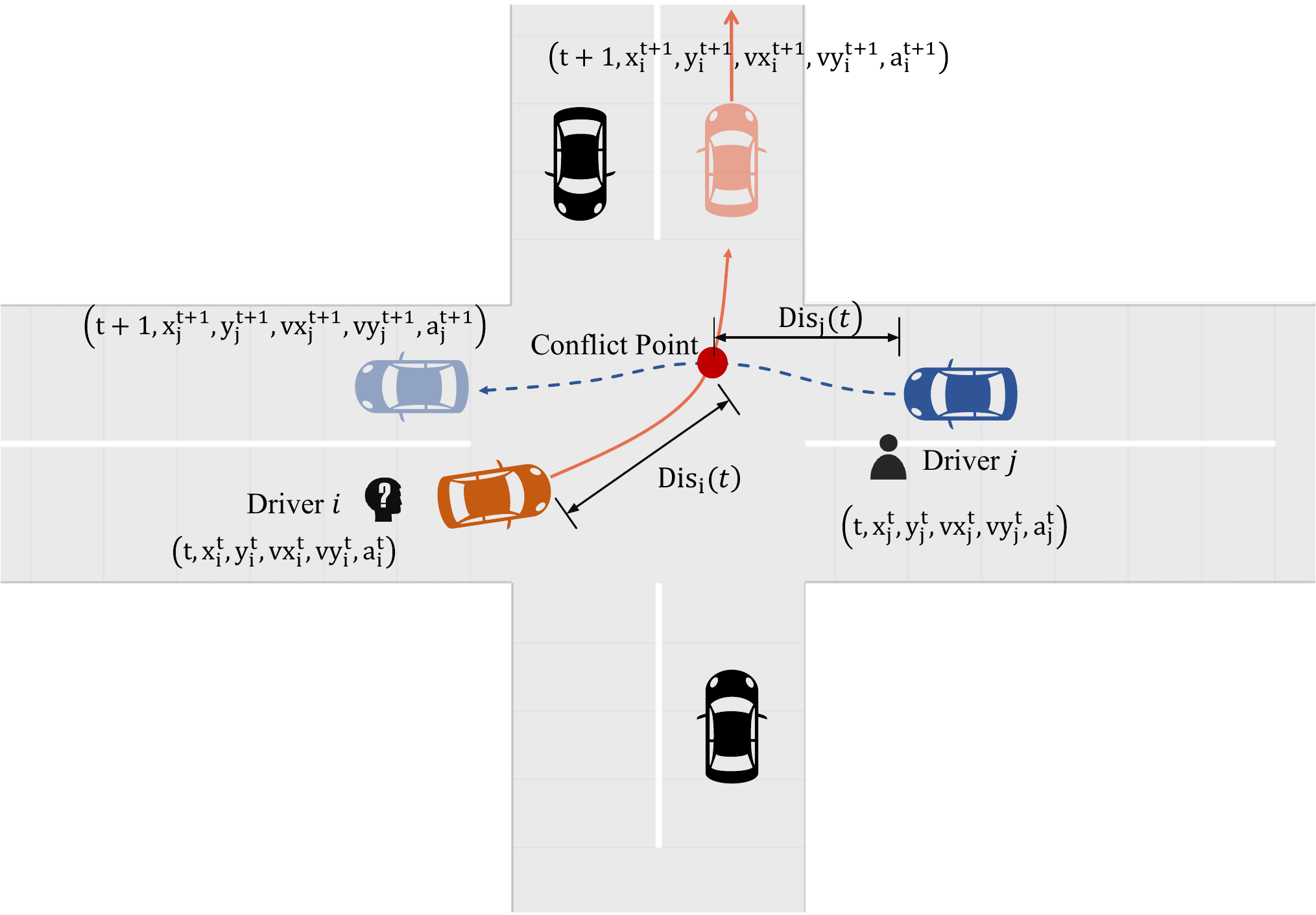}
    \caption{The interaction between one turn-left vehicle and one straight vehicle at the unprotected turning intersection. }
    \label{fig:left-turn}
\end{figure}

There are numerous social interaction scenarios and behaviors when drivers navigate the road, and making an unprotected left turn at an intersection is one of the most complex and challenging scenarios \citep{rahmati2021helping}. We use this scenario as a typical case to model and illustrate our evaluation framework in the subsequent sections.
When a driver makes an unprotected left turn at an intersection, potential conflicts with the straight vehicle in the opposite lane may arise, and interactions will occur in the process of determining the passing sequence. In our work, for simplicity, we mainly focus on the interaction between one unprotected left-turn driver and one straight driver at an intersection, as illustrated in Figure \ref{fig:left-turn}. 

\section{Design of The Evaluation Framework}
\label{section:design4}
In this section, we provide a detailed introduction to the evaluation framework, which includes Risk Perception Modeling, Interactive Process Modeling, and Interactive Ability Scoring.

\subsection{Risk Perception Modeling}
When turning left at an intersection, a driver must find a suitable gap for making the left turn while ensuring safety. To maintain driving safety, the driver will consider not only the current traffic environment risk but also the future risk associated with reaching the conflict point based on the motion state of the interacting object. Consequently, we propose the concept of comprehensive motion risk, defined as the sum of instantaneous state risk and future state risk. The risk field is utilized to quantify the perceived risk throughout the entire interaction process \citep{tan2021risk}.

\subsubsection{Instantaneous State Risk}
In accordance with the risk field theory, physical objects in the traffic environment are categorized into dynamic objects and static objects, with modeling calculations conducted separately \citep{wang2016driving}. Dynamic objects encompass entities with speeds greater than 0 or that may exceed 0 (motor vehicles, non-motor vehicles, pedestrians, etc.), while static objects include immobile obstacles and traffic lines. To quantify the instantaneous state risk of driver $d$, we reference the longitudinal driving risk model in \citep{tan2021risk}.

Within the Cartesian coordinate system, to quantify the risk of driver $d$, we consider all moving objects (vehicles) $i\ (i\in {1,2,…,n})$ in the vicinity of driver $d$. Let $[x(k),y(k)]$ denote the Cartesian coordinates of driver $d$ at the $k^{th}$ time interval. The risk factor in the x-axis direction is defined as:
\begin{equation}
    {\delta _{i}}(x(k)) = \frac{{{\beta _{x}}  \max (x(k) - (x_i + \frac{L_i}{2}  ,0))}}{{{\alpha _{x}}  {v_{i,x}}(t) + 1}}
    \label{eq:risk_now}
\end{equation}
where $\beta_{x}$ and $\alpha_{x}$  are the attenuation coefficients of the attenuation function of vehicle $i$ on distance and speed, respectively, which represent the effect of the speed and relative distance of moving object on the longitudinal risk, $v_{i,x}$ is the velocity component of the vehicle $i$ in the x-axis direction at $k^{th}$ interval, 
$x_i$ and $L_i$ are the x coordinate of the vehicle $i$ and the length of the vehicle $i$, respectively.

The risk function in the x direction is defined as:
\begin{equation}
    \textbf{R}_{i,longi} (x(k)) = \frac{1}{\delta_i (x(k))+1}
\end{equation}

Also for the y direction, the decay factor and hazard function are:
\begin{equation}
    \delta_i(y(k)) = \frac{\beta_y \max(y(k)-(y_i + \frac{W_i}{2}),0)}{\alpha_y v_{i,y}(t)+1}
\end{equation}
\begin{equation}
    \textbf{R}_{i,lateral}(y(k)) = \frac{1}{\delta _{i}(y(k))+1}
\end{equation}
where $W_i$ is the width of the vehicle $i$.

Considering the risk attenuation in the x and y directions comprehensively, we have:
\begin{equation}
    \delta_{i}(x(k),y(k)) = \sqrt{\delta_i^2(x(k))+\delta_i^2(y(k))}
\end{equation}

The risk function is defined as
\begin{equation}
    \textbf{R}_{i}(x(k),y(k)) = \frac{1}{\delta_{i}(x(k),y(k))+1}
\end{equation}

For all static objects $j(j \in \{ 1,2,...,m\})$, it is equivalent to a moving object with a speed of 0, and the corresponding risk function is derived by
\begin{equation}
    \textbf{R}_{lane,j} = a_j \frac{1}{\delta_{j}(x,y)+1}
\end{equation}
where $a_j$ denotes the maximum risk of the $j_{th}$ road line.

The risk values of all dynamic entities and static entities are superimposed to obtain the global instantaneous state risk field.


\begin{equation}
    \begin{aligned}
     & \textbf{R}_{now} (x(k),y(k)) = \sum_{i=1}^n \textbf{R}_{veh,i}(x(k),y(k)) 
     &  +  \sum_{j=1}^m \textbf{R}_{lane,j}(x(k),y(k))
    \end{aligned}
\end{equation}

\subsubsection{Future State Risk}
The EKF model is used to estimate the movement of the interactive vehicle for the next observation interval, which is a nonlinear version of the Kalman filter. EKF estimates a vehicle’s location with the state equation and observation equation first and then updates the vehicle’s location.

Let $\textbf{X} (k) = [x(k) \ y(k) \ v(k) \ \theta (k)]^T$, the estimated vehicle’s state at $k+1^{th}$ interval is described as follows.
\begin{equation}
    \textbf{X}_{k+1} = g(\textbf{X}(k),u(k))
\end{equation}
where $\textbf{X}_k$ represents the vehicle's state vector at $k^{th}$ interval, $[x(k) \ y(k)]$ denotes the Cartesian coordinates at $k^{th}$ interval, $v(k)$, $\theta(k)$  and $u(k)$ represent its speed, its heading angle and its actions at $k^{th}$ interval, respectively.

For motion prediction, the following unicycle model is used.
\begin{equation}
    \begin{bmatrix}
        x(k+1) \\ y(k+1) \\ v(k+1) \\ \theta(k+1) 
    \end{bmatrix}
    = g(\textbf{X}(k),u(k)) = 
    \begin{bmatrix}
        x(k)+v(k) \cos (\theta(k))\Delta t \\
        y(k)+v(k) \sin (\theta(k))\Delta t \\
        v(k) + a(k) \Delta t \\
        \theta(k) + \omega(k) \Delta t
    \end{bmatrix}
\end{equation}
where $a(k)$ and $\omega (k)$ represent the acceleration and yaw rate of the vehicle.

The EKF uses linear transformations to approximate nonlinear relationships. Specifically, the EKF linearizes the motion by taking partial derivatives of the process equations and using first-order Taylor expansions.

Firstly, the Jacobian matrix of $X(k)$ is computed.
\begin{equation}
    \textbf{J}_A(k)=
    \begin{bmatrix}
        \frac{\partial x(k)}{\partial x} \frac{\partial x(k)}{\partial y} 
        \frac{\partial x(k)}{\partial v} \frac{\partial x(k)}{\partial \theta} \\
        \frac{\partial y(k)}{\partial x} \frac{\partial y(k)}{\partial y} 
        \frac{\partial y(k)}{\partial v} \frac{\partial y(k)}{\partial \theta} \\
        \frac{\partial v(k)}{\partial x} \frac{\partial v(k)}{\partial y} 
        \frac{\partial v(k)}{\partial v} \frac{\partial v(k)}{\partial \theta} \\
        \frac{\partial \theta (k)}{\partial x} \frac{\partial \theta(k)}{\partial y} 
        \frac{\partial \theta (k)}{\partial v} \frac{\partial \theta (k)}{\partial \theta} \\
    \end{bmatrix}
\end{equation}

Then, the predicted state information of the vehicle is updated as follows \citep{schuhmann2011improving,schubert2011empirical}.
\begin{equation}
    \hat{\textbf{X}}(k+1 | k) = g(\textbf{X}(k),u(k))  \label{eq:trans}
\end{equation}
\begin{equation}
    \textbf{P}(k+1 | k) = \textbf{J}_A(k) \textbf{P}(k) \textbf{J}_A(k)^T + \textbf{Q}_k \label{eq:pro_trans}
\end{equation}
where $\hat{\textbf{X}}(k+1 | k)$ denotes the estimated value at $k+1^{th}$ interval with previous $k$ states, $\textbf{Q}(k)$ is the covariance matrix of noise in state $k$, $\textbf{P}(k+1 | k)$ denotes the predicted covariance matrix of the state of the vehicles. 

In our framework, the real position, velocity, and acceleration information of vehicles can be obtained from data and exchanged. The observation vector $\textbf{V}(k)$ can be written as
\begin{equation}
    \textbf{V}(k)=\textbf{H}(k)X(k)+\textbf{Z}(k) \label{eq:observation}
\end{equation}
where $\textbf{H}(k) = 
\begin{bmatrix}
    1 & 0 & 0 & 0 \\
    0 & 1 & 0 & 0 \\
\end{bmatrix}$, $\textbf{Z}(k)$ is the noise vector, which satisfies the normal distribution:$\textbf{V}(k) \sim N(0,\textbf{R}(k))$. $\textbf{R(k)}$ is the covariance matrix of noise at $k^{th}$ interval.

Besides, the Kalman gain is updated by
\begin{equation}
    \textbf{K}(k+1) = \textbf{P}(k+1 | k) H^T [\textbf{H}(k) \textbf{P}(k|k-1) H^T(k) + \textbf{R}(k)]^{-1}
\end{equation}

The optimal estimation  $\hat{\textbf{X}}(k+1)$ of current state k+1 is 
\begin{equation}
    \hat{\textbf{X}}(k+1) = \hat{\textbf{X}}(k+1 |k) + \textbf{K}(k+1) [Z(k+1)-H(k+1)\hat{\textbf{X}}(k+1|k)]
\end{equation}

The error equation is derived by
\begin{equation}
    \textbf{P}(k+1) = \textbf{P}(k+1|k) - \textbf{K}(k+1)\textbf{H}(k+1)\textbf{P}(k+1|k)
\end{equation}
where $\textbf{P}(k+1|k)$ is the vehicle’s predicted covariance matrix of state $k$.

Finally, the future state risk field is expressed as

\begin{equation}
    \begin{aligned}
     & \textbf{R}_{fe}(x(k),y(k)) = \sum_{i=1}^n \textbf{R}_{veh}(\hat{x}(k+1),\hat{y}(k+1)) \\
     &  +  \sum_{j=1}^m \textbf{R}_{lane}(x(k),y(k))
    \end{aligned}
\end{equation}

\subsubsection{Comprehensive Risk Perception}
The instantaneous state risk and future state risk are superimposed to obtain the quantification results of comprehensive risk perception from the global perspective.
\begin{equation}
    \textbf{R}_{all} =  \textbf{R}_{now}(x(k),y(k)) +  \textbf{R}_{fe}(x(k),y(k))
    \label{eq:comprehensive_risk}
\end{equation}

\subsection{Interactive Process Modeling}
The interaction between two drivers is a dynamic and complex behavioral process. Throughout this process, drivers dynamically adjust their behaviors based on other drivers' actions. Game theory can effectively describe this process, assuming that all participants act under entirely rational conditions, and an equilibrium solution can be obtained by solving the model. However, real-world drivers are not entirely rational, and their game performance will be worse than that of entirely rational players. As such, the game model solutions are defined as the exceptional interactive actions of rational individuals and are considered the criterion for evaluating real-world drivers' interaction abilities.

According to the game theory\citep{barron2013game}, there are at least three elements in the model: players, strategies, and profits. Assuming that there are n drivers interacting at an intersection, and the set of game players is $\mathcal{C}=\{C_L,C_S\}$, where $C_L$ is the set of the left-turn drivers, $C_S$ is the set of the straight drivers. The driver's strategy set is $S=\{s_1,s_2,s_3\}$, where $s_1$, $s_2$ and $s_3$ represent acceleration , uniform speed and deceleration, respectively. The position and speed states of driver $i$ at $k^{th}$ time interval are determined by his historical acceleration set $\mathcal{A}_i^{k-1}$, where $\mathcal{A}_i^{k-1}=(a_i^0,a_i^1,…,a_i^{k-1})(1 \leq k \leq T-1)$, and the total benefits of driver $i$  $u_i^k$ during game process is also determined by the same set $\mathcal{A}_i^{k-1}$, remember to $u_i^k= \Phi (a_i^0,a_i^1,…,a_i^{k-1}) = \Phi (\mathcal{A}_{k-1}^i)$.

Let $v_i(k-1)$ denote velocity of player $i$ at $k-1^{th}$ time interval. When taking action at $k^{th}$ time interval, the speed during interval $[k-1,k)$ is updated as
\begin{equation}
    v_i (k)=v_i (k-1)+a_i (k) \Delta t
\end{equation}
where $a_i (k)$ represents corresponding acceleration or deceleration action player $i$ takes from strategies at $k^{th}$ time interval.

Then the travel distance and estimated travel time of player $i$ are calculated by the following function.
\begin{equation}
    s_i (k)=s_i (k-1)+v_i (k)\Delta t
\end{equation}
\begin{equation}
    t_i = \frac{L_i - s_i(k)}{v_i(k) + \epsilon}
\end{equation}
where $s_i (k)$ denotes the travel distance of player $i$ at $k^{th}$ time interval, $s_i (0)=0$, $L_i$ is the distance to the conflict point of player $i$ at the intersection, and $\epsilon$ is a tiny positive constant which can ensure the function to be well-defined.

The type of game model and the setting of the player payoff function will significantly affect the model results. From an individual driver's perspective, safety and efficiency are the two most important factors influencing decision-making, so safety and efficiency payoffs are considered for players. To evaluate drivers' performance in various aspects, three types of rational individuals with different driving preferences are defined by adjusting players' payoff functions.

To evaluate driving safety and travel efficiency, three preferred rational individuals, i.e., safety-first, efficiency-first, and comprehensive-type, are defined.

\textbf{Safety-first Rational Person}: The driver who only considers safety risk when making decisions. The game benefit function of the player is written as
\begin{equation}
    us_i^k (\alpha_i )=1-R_i(k)
    \label{eq:safety-first}
\end{equation}
where $R_i(k)$ represents the perceived comprehensive risk derived from our risk perception modeling, as calculated in \ref{eq:comprehensive_risk}.

\textbf{Efficiency-first Rational Person}: The driver who only takes the prevailing efficiency as the decision-making criterion. The game benefit function of the player is expressed as
\begin{equation}
    ue_i^k (\alpha_i )=E_i(k)
    \label{eq:efficiency-first}
\end{equation}
where $E_i(k)$ is efficiency term and denoted by the velocity of the vehicles. The higher the velocity, the greater the efficiency.

\textbf{Comprehensive-type Rational Person}: The driver who considers safety and efficiency factors to make comprehensive decisions when interacting, and his(her) benefit function is derived by
\begin{equation}
    ub_i^k (\alpha_i)=m_i^k E_i(k)+n_i^k (1-R_i(k))
    \label{eq:Comprehensive-type}
\end{equation}
where $R_i(k)$ is the comprehensive risk of driver $i$ perceived at $k^{th}$ interval, as formulated in the Eq. \ref{eq:comprehensive_risk}, $E_i^k$ is the efficient benefit of driver $i$, which is defined as the speed of the vehicle per the unit time, $\alpha ^k _i$ is the specific action performed by driver $i$, $m_i^k$ and $m_i^k$ respectively represent the driver’s weight coefficient for efficiency and safety risk perception, $m_i^k \geq 0$, $n_i^k \geq 0$, $m_i^k+n_i^k=1$. It should be noted that safety-priority and efficiency-priority drivers are only ideal boundaries and do not exist in reality.

Game theory has traditionally been divided into two categories: non-cooperative games and cooperative games \citep{barron2013game}. In non-cooperative games, all players make decisions independently and only consider maximizing their own payoffs. In cooperative games, higher overall payoffs can be obtained by reaching a consensus among individuals and forming a coalition \citep{yang2018cooperative}. Non-cooperative and cooperative games represent two kinds of interactive thinking.

In reality, drivers have different degrees of cooperative and competitive tendencies, which can be quantified through the Social Value Orientation index \citep{wang2022social}. In our paper, two boundary cases are defined: one is the completely selfish rational individual who will only consider their own objectives and maximize their utility when passing through the intersection, leading to a non-cooperative interaction game; the other is the total-altruism rational individual who will completely prioritize the efficiency of others. When two completely altruistic individuals appear, both parties will take action with the goal of maximizing the system's total efficiency, generating a cooperative interaction game. Therefore, to evaluate the competitiveness and cooperation of driver interaction, two types of game models are set, i.e., non-cooperative games and cooperative games.

\subsubsection{Non-cooperative Game}
In the non-cooperative game, players will determine their strategies independently and aim to maximize their payoff, which leads to competition and can be characterized as a Nash game.

For player $i$ , whose rational style is denoted as $\mu (\mu \in {\mu s,\mu e,\mu b})$, the utility at every time interval is $u_{t_i}$. $\mu s,\mu e,\mu b$ are defined in formula \ref{eq:safety-first},\ref{eq:efficiency-first},\ref{eq:Comprehensive-type} respectively. Therefore, at $k^{th}$ interval, the optimization objective function of the player $i$ can be characterized as

\begin{equation}
    \textbf{a}_i^{k \ast}  = \arg \max \ u_{t_i}^{\ast} (k) = \arg \max \Phi (\mathcal{A}_i^{k-1},a^i_k)
\end{equation}

s.t.
\begin{equation}
    \| t_i - t_{i^-} \| \geq t_{avoid}, {\forall} i \in \{C_L,C_S\}, i^- \in \{C_L,C_S\} \backslash \{i\}  \label{game_s_1}
\end{equation}
\begin{equation}
    0 \leq v_i(k) \leq v_{max}, {\forall} i \in \{C_L,C_S\} \label{game_s_2}
\end{equation}
\begin{equation}
    \| a_i(k) \| \leq a_{max} , {\forall} i \in \{C_L,C_S\} \label{game_s_3}
\end{equation}
where $t_{avoid}$ is the minimum time to avoid the collision, which is related to the intersection size and speed of the players,  $v_{max}$ and $a_{max}$ are the maximum permitted speed under traffic rules and maximum comfortable acceleration driver can choose.

The Lemke-Howson Algorithm (L-HA) is used to solve the Nash equilibrium of the non-cooperative model\cite{shapley1974note}, which is defined as the competitive result of players.
\subsubsection{Cooperative Game}

When two completely altruistic players interact, two parties will take action to improve the system's efficiency, which leads to a cooperation game. Different from the non-cooperative game, in the cooperative game framework, drivers will reach cooperative consensus and form coalitions to gain higher overall benefits.

For a cooperative coalition, there exists a set $N$ (containing $n$ players) and eigenfunction $\mathcal{V}$ that maps a subset of players to the real numbers, $\mathcal{V}:2^n \to R$. $\mathcal{V}(S)$ represents the total payoff obtained by all members of coalition $S$ through cooperation. In a given coalition game $(\mathcal{V},N)$, the benefit of any player $i$ in the set at time step $k$ is calculated by Shapely theory as follows:
\begin{equation}
    u^k_i = \sum_{S \subseteq N} \sum_{\forall i \in N} \frac{(\lvert S \rvert -1)! (n-\lvert S \rvert)!}{n!} [\mathcal{V}^k(S)-\mathcal{V}^k (S-\{i\})]
\end{equation}
where $\lvert S \rvert$ denotes the number of members in the coalition $S$.
According to the definition of eigenfunction $\mathcal{V}$ and game profit, it yields that
\begin{equation}
    \label{eq:cases}
    \begin{cases}
    \mathcal{V}^k[\emptyset] = 0 \\

    \mathcal{V}^k[\{a,b\}] = \big(m_a^k E_a (k)+n_a^k (1-R_a (k))+ (m_b^k E_b (k) \\
     +n_b^k (1-R_b (k) \big) \\
    \vdots \\
    \mathcal{V}^k[N] = \big(m_1^k E_1 (k)+n_1^k (1-R_1 (k))+(m_2^k E_2 (k) \\
    +n_2^k (1-R_2 (k))+ \cdots +(m_N^k E_N (k)+n_N^k (1-R_N (k) \big) \\
    \end{cases}
\end{equation}
\begin{equation}
    \mathcal{V} ^k (N)=\sum_{\forall i \in N} u_i^k 
\end{equation}

The reason that Eq. \ref{eq:cases} always holds is presented as follows.

\begin{equation}
    \begin{aligned}
     & \sum_{i \in N}u^k_i = \frac{1}{\lvert N \rvert !} \sum_{S \subseteq N} \sum_{\forall i \in N} \big(\mathcal{V}^k(S) - \mathcal{V}^k(S-{i})\big)\\
     &  = \frac{1}{\lvert N \rvert !} \sum_{S \subseteq N} \mathcal{V}(N) = \frac{1}{\lvert N \rvert !} \lvert N! \rvert \mathcal{V}(N) = \mathcal{V}(N) \\  
    \end{aligned}
\end{equation}

In the cooperative game, solving the optimal action sequence at each time step is equivalent to an optimization problem, and its optimization objective is defined by
\begin{equation}
    (a_1^{k \ast},a_2^{k \ast},\cdots,a_n^{k \ast}) = \arg \max \sum_{i=1}^n \Phi^k_i(\mathcal{A}^{k-1}_i,a^k_i)
\end{equation}
s.t.
\begin{equation}
    \| t_i - t_{i^-} \| \geq t_{avoid}, {\forall} i \in \{C_L,C_S\}, i^- \in \{C_L,C_S\} \backslash \{i\}  \label{game_s_1}
\end{equation}
\begin{equation}
    0 \leq v_i(k) \leq v_{max}, {\forall} i \in \{C_L,C_S\} \label{game_s_2}
\end{equation}
\begin{equation}
    \| a_i(k) \| \leq a_{max} , {\forall} i \in \{C_L,C_S\} \label{game_s_3}
\end{equation}
\begin{equation}
    \mathcal{V}(S) + \mathcal{V}(T) \leq \mathcal{S+T}, \forall S,T \in N, S \cap T = \emptyset
\end{equation}

The Genetic Algorithm (GA) is used to solve the Pareto optimal issue, which is defined as the cooperative result of the interacting objects\citep{yang2018cooperative}. The payoff $u_i^k$ of each player in the coalition can be described as follows.
\begin{equation}
    \Omega ^k = [\Phi_1^k,\Phi_2^k,\cdots,\Phi_n^k]
\end{equation}

Finally, the optimal action solution results of all players are obtained.
\begin{equation}
    \begin{aligned}
    & \mathcal{A}^{k \ast} = [\mathcal{A}^k_1,\mathcal{A}^k_2,\cdots,\mathcal{A}^k_n] \\
    & =  [\mathcal{A}^{k-1}_1 \cap a^{k \ast}_1,\mathcal{A}^{k-1}_2 \cap a^{k \ast}_2,\cdots,\mathcal{A}^{k-1}_1 \cap a^{k \ast}_1]
    \end{aligned}
\end{equation}

According to the game strategies executed by two players at $k^{th}$ interval, the motion states of players at $k+1^{th}$ are calculated. Then the game strategy is selected again at $k+1^{th}$ step. The whole modeling and solving process is shown in Algorithm~\ref{alg:algorithm 1}. 

\subsection{Interactive Ability Scoring}
The rational individual's actions serve as the evaluation benchmark for assessing interaction ability. We measure the interaction performance of drivers by comparing their actual actions to the idealized rational individual acceleration sequences. Currently, evaluations of driver decision-making primarily utilize trajectory information, leveraging metrics such as KL divergence, MSE, and mADE \citep{demetriou2020generation}.

However, relying solely on trajectory information for interaction evaluation falls short, as it omits temporal dynamics and may not fully capture the nuances of a driver’s reaction speed, intensity, and action preferences. Furthermore, the existence of multiple optimal trajectories—represented as various equilibrium solutions in the game model—complicates the assessment.

In contrast, acceleration data provide a richer depiction of driver behavior. For instance, more adept drivers typically exhibit smoother acceleration and deceleration patterns, anticipate upcoming interactions, and adjust their speed preemptively, all discernible from acceleration sequences. These sequences, represented as N-dimensional time series vectors, are utilized to enhance the evaluation of driver interactions.

For a specific driving scene, the actual acceleration sequence of interactive object $i$ is expressed as
\begin{equation}
    \mathcal{A}^{ \ast}_{re} = [a_1,a_2,a_3,...,a_n]
    \label{eq10}
\end{equation}

The excellent acceleration sequence for the rational person is described as 
\begin{equation}
    \mathcal{A}^{ \ast}_{so} = [\hat{a_1},\hat{a_2},\hat{a_3},...,\hat{a_n}]
\end{equation}

Thus, the ability evaluation problem is transformed into a vector similarity evaluation problem. Common similarity evaluation indices include Manhattan distance and Euclidean distance \citep{magdy2015review}, but these methods overlook many important vector features. An improved evaluation method based on morphological similarity distance is proposed, considering three aspects: Euclidean distance for vector distance measurement, cosine similarity for vector direction evaluation, and morphological similarity evaluation of time series data, offering a more comprehensive assessment than a single index.

The improved morphological similarity distance between vectors $\mathcal{A}^{n \ast}_{re}$ and $\mathcal{A}^{n \ast}_{so}$ can be defined as follows.
\begin{equation}
    D_{MSD,i}(\mathcal{A}^{\ast}_{re},\mathcal{A}^{\ast}_{so}) = D_{ED,i}(\mathcal{A}^{ \ast}_{re},\mathcal{A}^{ \ast}_{so}) * \big(2-\frac{ASD_i}{SAD_i}\big) 
\end{equation}
\begin{equation}
    ASD_i = \lvert \sum_{i=1}^n {(a_k - \hat{a}_k)} \rvert
\end{equation}
\begin{equation}
    SAD_i = \sum_{i=1}^n  \lvert a_i - \hat{a}_i \rvert
\end{equation}
where $D_{ED,i}$ denotes the Euclidean distance between $\mathcal{A}^{ \ast}_{re}$ and $\mathcal{A}^{ \ast}_{so}$, $SAD$ (Sum of Absolute Differences) is the Manhattan distance between $\mathcal{A}^{ \ast}_{re}$ and $\mathcal{A}^{ \ast}_{so}$, $ASD$ (Absolute Sum of Differences) is the absolute value of the sum of dimensional differences between $\mathcal{A}^{ \ast}_{re}$ and $\mathcal{A}^{ \ast}_{so}$.

The cosine similarity between $\mathcal{A}^{ \ast}_{re}$ and $\mathcal{A}^{ \ast}_{so}$ is defined as follows.
\begin{equation}
    Co_i  = \frac{\mathcal{A}^{ \ast}_{so} \mathcal{A}^{ \ast}_{re}}{\| \mathcal{A}^{ \ast}_{re} \| \| \mathcal{A}^{ \ast}_{so} \|} = \frac{\sum_{i=1}^n a_1 \hat{a_1}}{\sqrt{\sum_{i=1}^n a_i^2}+\sqrt{\sum_{i=1}^n \hat{a_i}^2}} \label{eq11}
\end{equation}

And then the interactive ability score is defined by
\begin{equation}
    S_i = P_{MASD,i} = \frac{Co_i}{1+D_{MSD,i}}
\end{equation}
where $S_i\in [-1,1]$.

The interaction ability score results are categorized into five levels, with level I representing the best interaction ability and level IV representing the worst interaction performance. In specific evaluations, the one-way ability or comprehensive ability of interactive vehicles can be scored and rated, allowing for a horizontal comparison.

\begin{algorithm} [!htbp]
    \caption{Game-based Interaction Process.}
    \label{alg:algorithm 1}
    \LinesNumbered 
    \KwIn{Initial states $X_i(k)$ of player $i \ (i \in \mathcal{C})$  }
    \KwOut{The action sequence sets $\mathcal{A}^*$ }
    Initialize time interval $k \gets 0$;

    Calculate the conflict point of two players $CP_{target}$;
    
    Choose the player style $\mu (\mu  \in \{\mu s,\mu e,\mu b\}) $and game type $\tau(\tau \in \{ NC,C \}) $; 
    
    \While{$k \geq 0$}{
        \For{$i \in \mathcal{C}$}{
        \If{player $i$ have passed the point $CP_{target}$}
        {Break}
        
        \textbf{Update} the total benefit $u_i (k)$ of player $i$:
        
        $u_i(k) \gets R_i(k) +E_i(k)$;
        
        \textbf{Solve} for equilibrium and get strategy $\alpha_i(k)$: 

        \eIf{$\tau = NC$}{
			$\alpha_i(k) \gets $ Solved by L-HA;
		}{
			$\alpha_i(k) \gets $ Solved by GA \;
		}
        
        \textbf{Update} the state of the player: 
        
        $v_i (k)=v_i (k-1)+\alpha _i (k)\Delta t$;
        
        $s_i (k)=s_i (k-1)+v_i (k) \Delta t$;
        
        \textbf{Update} $\mathcal{A}_i^k \gets \mathcal{A}_i^{k-1} \cap \alpha_i(k)$;
        
        $k \gets k+1$;
        
        }
    $\mathcal{A}^* \gets \mathcal{A}_i^k $;
    }
    
\end{algorithm}

\section{Experiments and Results}
\label{section:exper5}
In this section, the process of parameter calibration for the model is first described. Then the differences in driver ability in two datasets from three aspects are analyzed, and several cases are selected and mined to observe the drivers’ reaction during the interaction. Finally, we discuss the differences between our model and another baseline method, and the comparison results reflect the advantages of our model.

\begin{figure}[!htbp]
    \centering
    \includegraphics[width=1 \textwidth]{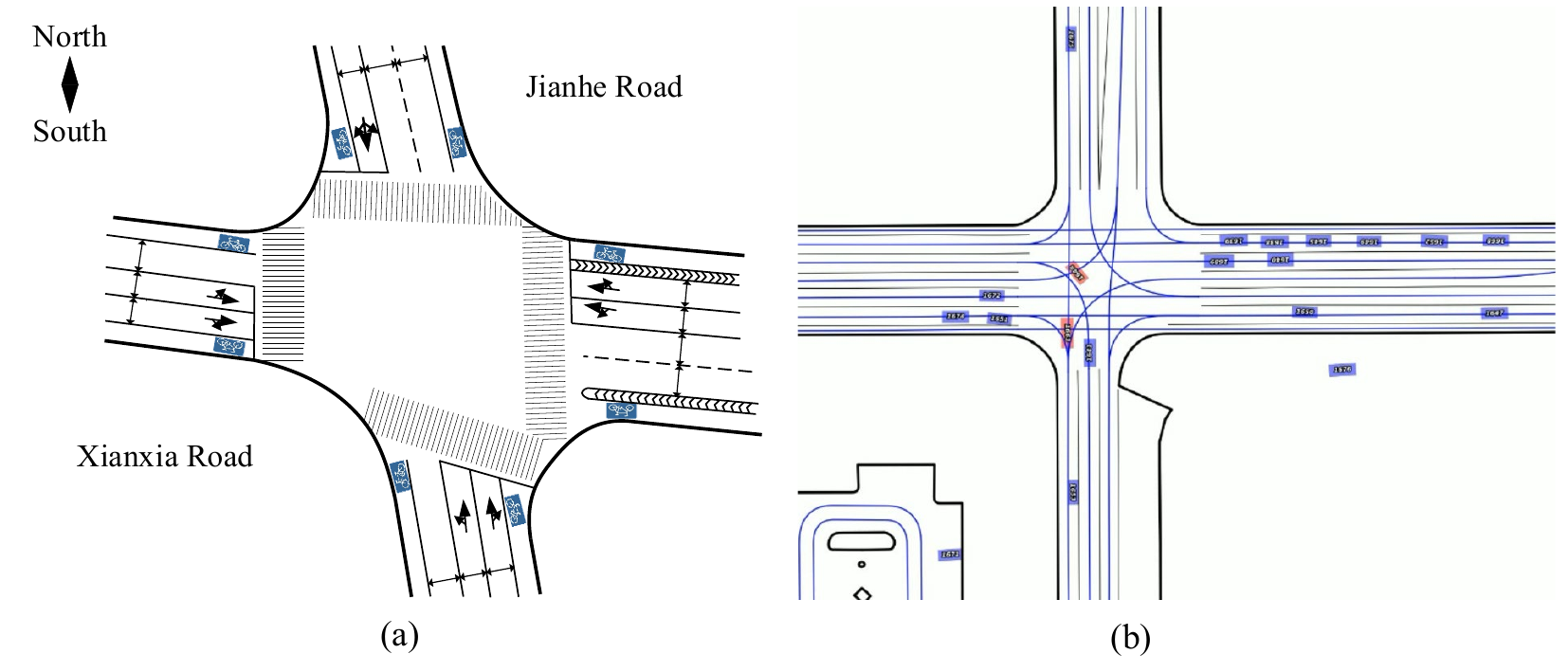}
    \caption{The left turn events from two datasets: (a) the Xianxia-Jianhe Intersection; (b) the intersection from Waymo Left-Turn Motion Dataset. 
}
    \label{fig:intersection_description}
\end{figure}

\subsection{Data and Scenario}
We process two datasets for trial evaluation and comparative analysis, as shown in Figure \ref{fig:intersection_description}. 

The first dataset employed is the Xianxia-Jianhe Dataset, sourced from the Jianhe-Xianxia Road intersection in Shanghai, China \citep{ma2017two}. This intersection is a typical two-phase junction where vehicles making left turns frequently encounter both motorized and non-motorized vehicles traveling straight.
To extract empirical data, we used high-precision video processing software. A camera was strategically positioned on a high building at the northeast corner of the intersection under clear weather conditions. It recorded traffic activities during the peak period from 4 PM to 5:40 PM. Our analysis primarily focused on vehicles executing left turns from the east approach, engaging predominantly with oncoming traffic. From the comprehensive trajectory data analyzed, we isolated 54 specific unprotected left-turn events that featured complete, one-on-one interactions, excluding any multi-vehicle interaction scenarios.

The second dataset utilized is the Waymo Left-Turn Motion Dataset, derived from the Waymo Open Motion Dataset, released in March 2021. This dataset features a vast array of interactive behaviors across various scenarios such as car following, lane changing, and intersection turning in the United States \citep{ettinger2021large}. The data processing involves three key stages:
(1) Data Preprocessing: We established a TensorFlow deep learning environment to parse the Scenario format dataset, extracting trajectory, motion state, and map details from each scenario, and transformed these into a two-dimensional data table. 
(2) Map Data Processing: Using map data, we identified intersection interaction scenarios and extracted the topological relationships of all lanes, noting the location, dimensions, angles, and structures of the intersections.
(3) Motion Object Analysis and Interaction Behavior Extraction: We analyzed the motion state and position of all moving objects, along with their environmental context, such as lane occupancy and proximity to intersections. We developed criteria based on distance and angle to identify vehicles engaged in unprotected left turns and their interaction dynamics. 
From the processed Waymo Open Motion Dataset, we analyzed trajectory segments to identify 794 left-turn interaction scenarios, specifically targeting complex situations where vehicles make unprotected left turns across opposing traffic lanes. After additional filtering for intersection scenario requirements, one-on-one interactions, and complete trajectory data, we isolated 192 specific unprotected left-turn events.

In the experiment part, the model parameters are first calibrated with real-world datasets. The differences in drivers' safety, efficiency, and interactive sociality during interaction from the two datasets are evaluated. Besides, a group of control experiments is set up. The risk field model of game safety profit is replaced by a simple prediction index based on PET to observe the impact of the risk field model on the evaluation effect.

\subsection{Parameters Calibration for Game Model}
The model's accuracy and the evaluation's reliability depend on the calibration of the model parameters. In this subsection, we detail the process and results of parameter calibration.

Parameter calibration can be considered an optimization problem. The optimization objective is to search for the best set of parameters, minimizing the gap between drivers' actions and game model solutions. Formally, the optimal objective can be expressed as:

\begin{equation}
    F = \frac{1}{2} \sum_{i \in \{ C_L,C_S \}} \frac{1}{N_i} \sum^{N_i}_{j=1} \sum^T_{t=1}( a^t_{ij}   -  \hat{a}^t_{ij}  )^2
\end{equation}
where $a_{ij}^t$ and $\hat{a}^t_{ij}$ denote the acceleration of the real-world driver and the game model at the $t^{th}$ timestamps, $N_i$ denotes the number of the type $i$ drivers used to be calibrated.

The parameters to be optimized are selected from the part of the risk field model most affected by the initial parameter settings. They include (1) weighting factors of instantaneous state risk and future state risk $w_n$; (2) the attenuation coefficients of the attenuation function for speed on the x-axis and y-axis $\alpha_x$ and $\alpha_y$; (3) the attenuation coefficients of the attenuation function for distance on the x-axis and y-axis $\beta_x$ and $\beta_y$. The search interval for all parameters is 0.05.

The genetic algorithm (GA) is used to solve the optimization problem. Since the optimization goal of the genetic algorithm is to maximize the fitness function, we set the fitness function in the algorithm as:
\begin{equation}
    F^{'} = \frac{1}{F}
\end{equation}

We set different population sizes $P (P \in { 10,20 })$ for the GA and different numbers $N (N \in { 1,5,10,30})$ of drivers used for calibration to observe the performance of the algorithm. The other hyperparameters of the GA are shown in Table \ref{tab:Hyperparameters for GA}.

\begin{table}  
    \centering
    \caption{The hyperparameters of the genetic algorithm.}
    \setlength{\tabcolsep}{12mm}{  
    \begin{tabular}{c c}
        \toprule
        Hyperparameter & Value \\
        \midrule
        DNA Size & 10 \\
        Crossover Rate & 0.6 \\
        Mutation Rate & 0.01 \\
        N Generations & 50 \\
        \bottomrule
    \end{tabular}
    }
    \label{tab:Hyperparameters for GA}
\end{table}

\begin{figure}  
    \centering
    \includegraphics[width=0.9 \textwidth]{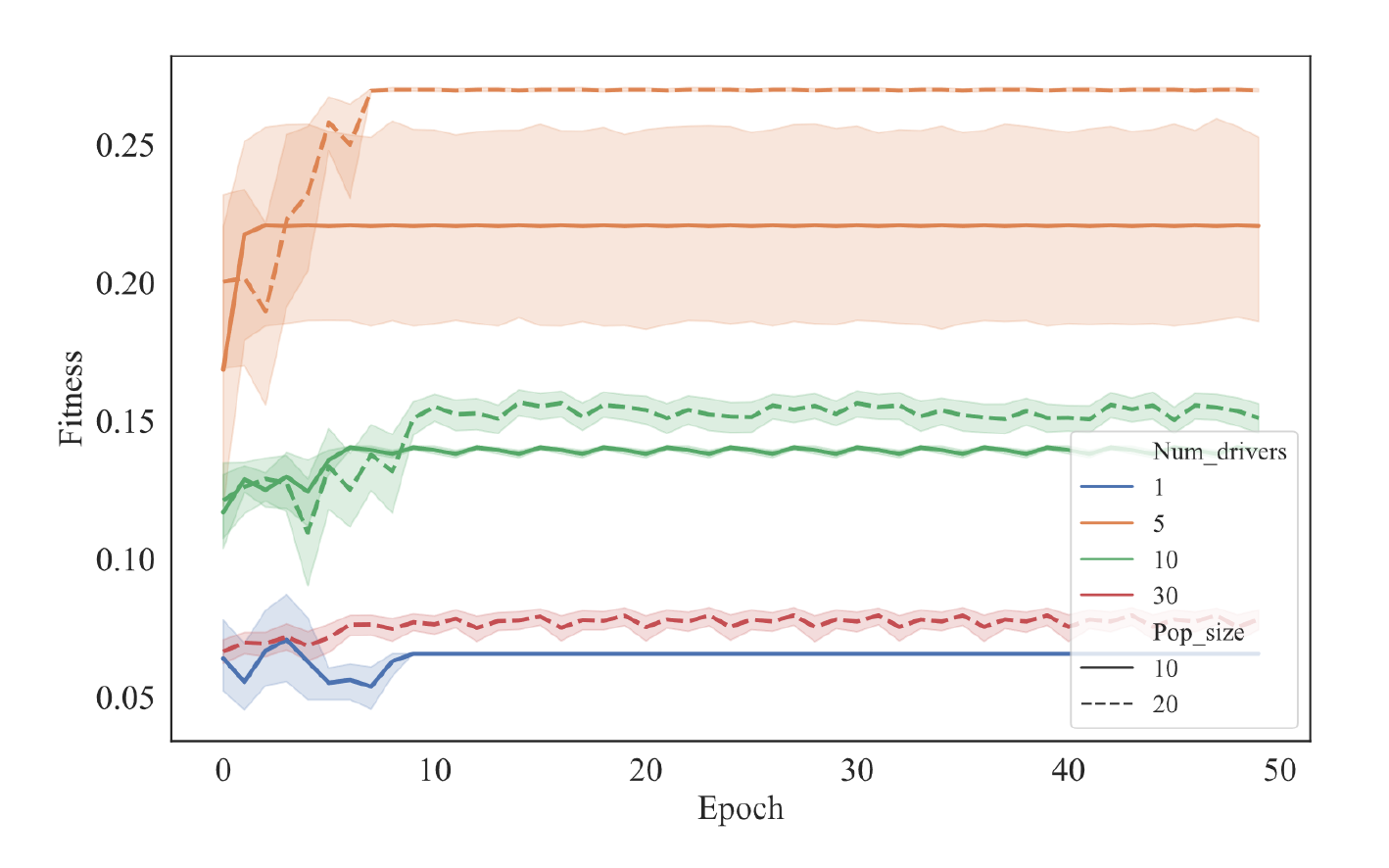}
    \caption{The fitness of the objective function with different hyperparameters during the optimization process.}
    \label{fig:para_calibration}
\end{figure}

The fitness of the objective function with different experimental conditions during the search process is exhibited in Figure \ref{fig:para_calibration}. The search for all conditions eventually converges within 50 iterations. Moreover, the number of drivers and the population size for GA significantly affect the solving results. Too few or too many drivers make calibration more difficult. Compared with the numbers 1, 10, and 30, calibrating with five drivers is more appropriate. Additionally, a larger population size produces better calibration performance, contrasting population sizes 10 and 20.

And the final calibration results are:
\begin{equation}
    [w_n, \alpha_x,\alpha_y,\beta_x, \beta_y] = [0.3157,0.1053,0.4737,0.8421,0.8947] \nonumber
\end{equation}

\subsection{Results and Analysis}
In this subsection, we analyze the differences in drivers' interactive abilities between two datasets based on the calibrated parameters.  Three kinds of characteristics are evaluated and analyzed, including safety, efficiency, and interactive sociality. The ability scores for drivers across all interaction scenarios are calculated, providing a statistical description of driving abilities and styles from different countries and cultures. We calculate the average ability scores for drivers and analyze the distribution of scores. Furthermore, we examine typical cases using both non-cooperative and cooperative game solvers.

\subsubsection{Safety and Efficiency Ability}
\begin{figure}  
    \centering
    \includegraphics[width=0.9 \textwidth]{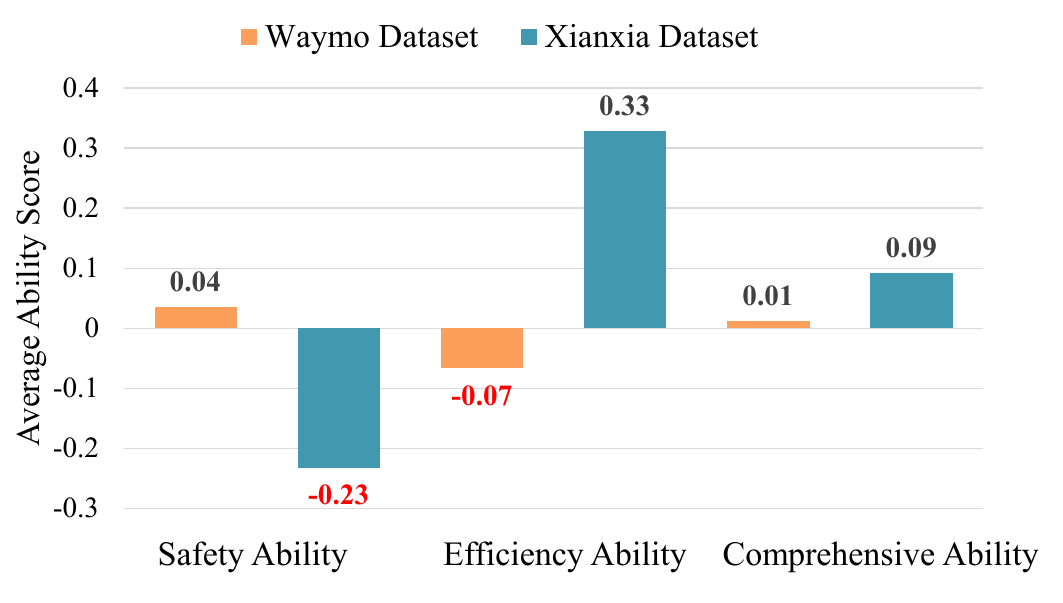}
    \caption{Drivers’ average ability score from two datasets.}
    \label{fig:ave_score}
\end{figure}

In the non-cooperative game framework, the drivers' performance in three criteria, including safety, efficiency, and comprehensive ability of safety and efficiency, are evaluated. Figure \ref{fig:ave_score} shows the average scores for left-turn drivers in both datasets.

Regarding comprehensive ability, Waymo drivers (from the United States) have an average score of 0.01, while Xianxia drivers (from China) have an average score of 0.09, indicating that Xianxia drivers demonstrate better comprehensive left-turn abilities. For safety performance, Waymo drivers score an average of 0.04, while Xianxia drivers score an average of -0.23, indicating that Waymo drivers perform significantly better in terms of safety. In terms of efficiency performance, Waymo drivers score an average of -0.07, while Xianxia drivers score an average of -0.33, suggesting that Xianxia drivers are more efficient.

\begin{figure}  
    \centering
    \includegraphics[width=1 \textwidth]{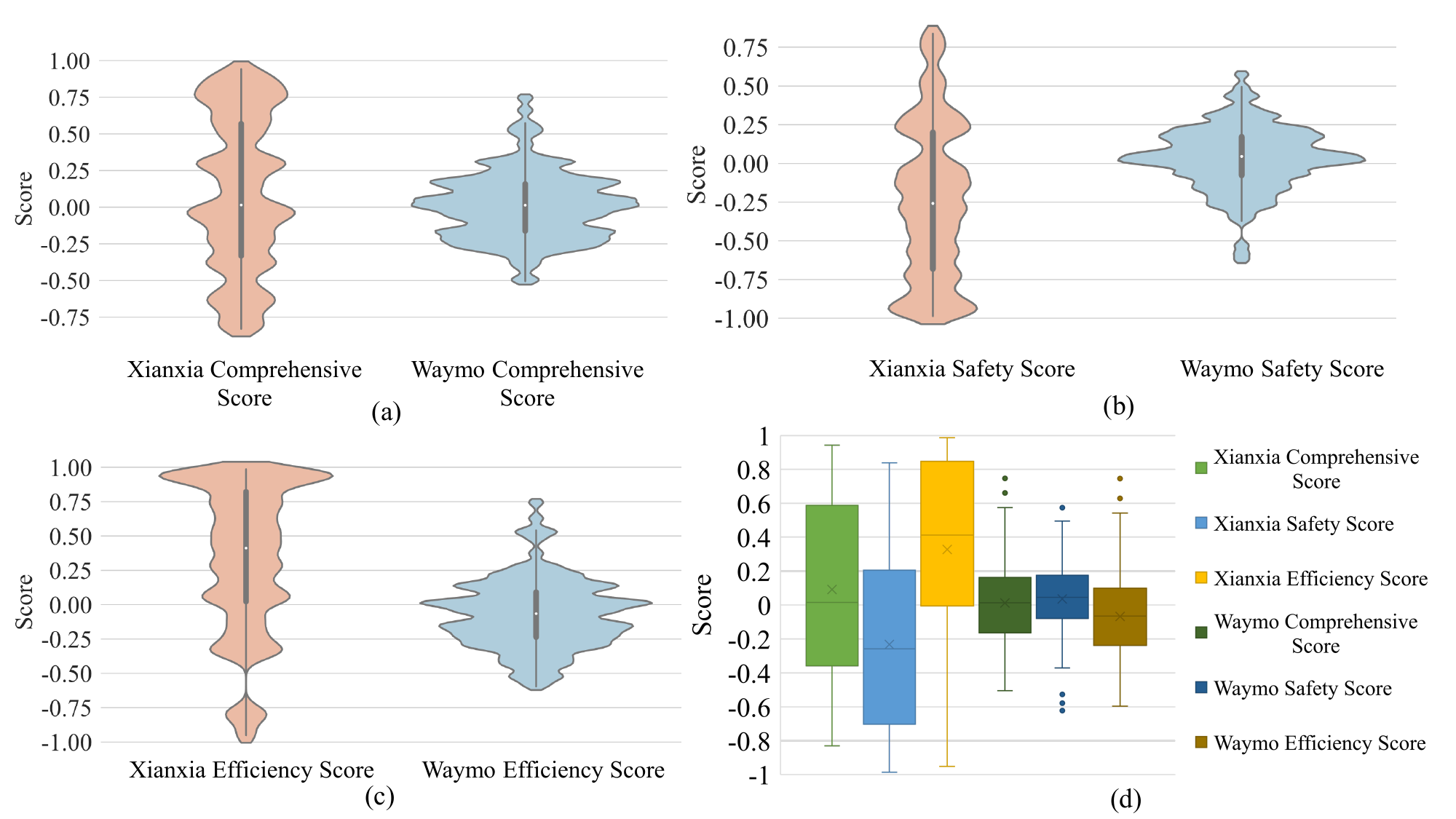}
    \caption{Distribution of driver ability scores: (a) Driver comprehensive ability score violin chart; (b) Driver safety score violin chart; (c) Driver efficiency score violin chart; (d) Box plots of mean scores for the two datasets.}
    \label{fig:violin_box}
\end{figure}

\begin{figure}[!htbp]
    \centering
    \includegraphics[width=1 \textwidth]{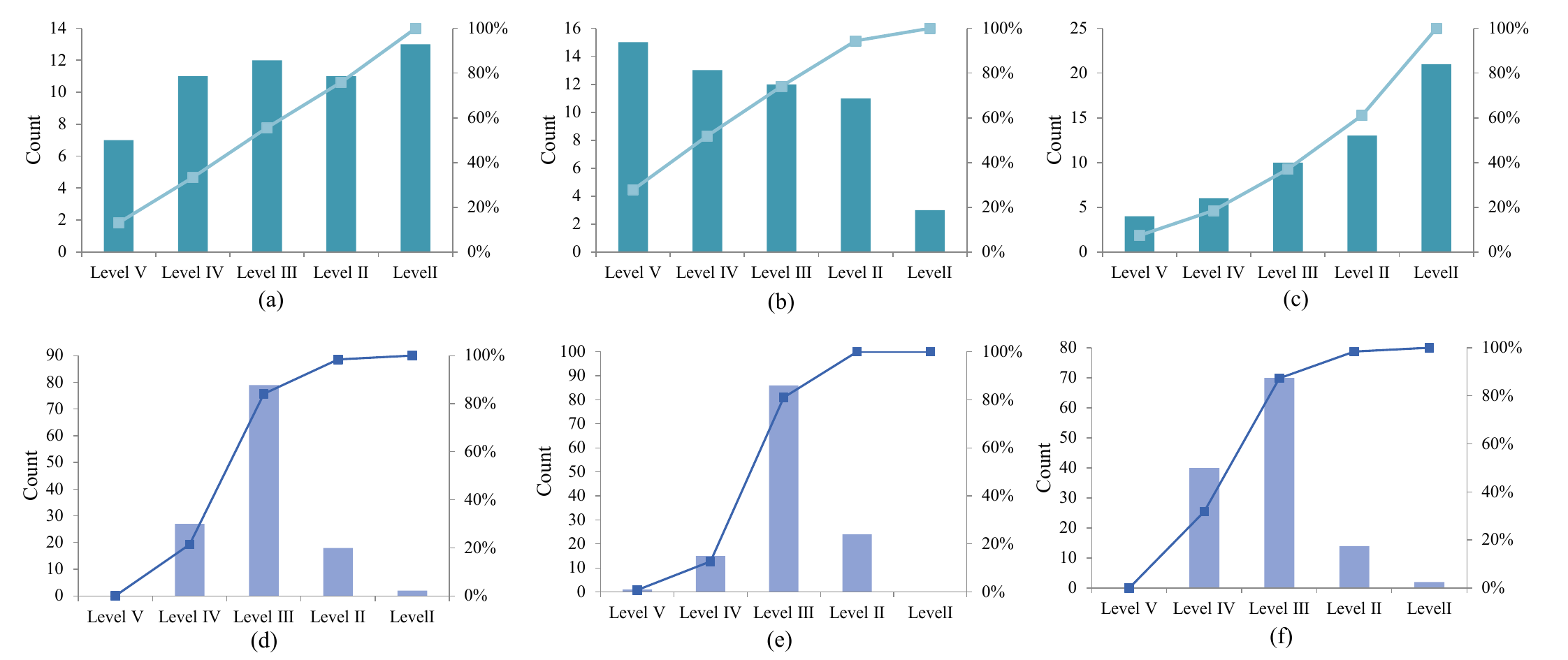}
    \caption{Driving ability class frequency map: (a) Xianxia dataset - Comprehensive capability; (b) Xianxia Dataset - Safety capability; (c) Xianxia dataset - Efficiency capability; (d) Waymo dataset - Comprehensive capabilities; (e) Waymo dataset - Safety capabilities; (f) Waymo dataset - Efficiency capability.
}
    \label{fig:ability_level}
\end{figure}

Figure \ref{fig:violin_box} illustrates the differences in score distributions for both driver types across the three criteria, and Figure \ref{fig:ability_level} presents the distribution of ability levels. Overall, the results indicate a significant difference in driving preferences between Xianxia and Waymo drivers when making unprotected left turns. Xianxia drivers prioritize efficiency over safety, while Waymo drivers are more conservative and place greater emphasis on safety.

\subsubsection{Interactive Sociality}
\begin{figure}  
    \centering
    \includegraphics[width=1\textwidth]{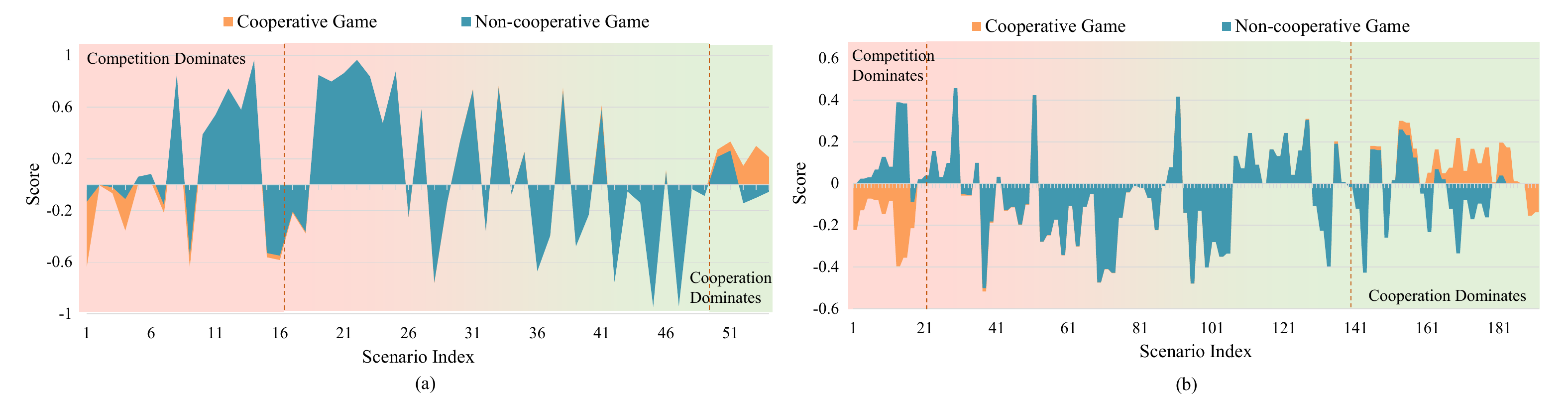}
    \caption{The competition and cooperation ability of drivers from two datasets: (a) The drivers from Xianxia Dataset; (b) The drivers from Waymo Dataset.}
    \label{fig:competition_cooperative}
\end{figure}

As previously mentioned, we employ levels of competition and cooperation to depict interactive sociality. We design non-cooperative and cooperative game frameworks to evaluate drivers' performance in terms of cooperation and competition. The Nash equilibrium in the non-cooperative game and the Pareto optimal solution in the cooperative game are considered as the ability measuring standards. Figure \ref{fig:competition_cooperative} shows the drivers' comprehensive ability results.

In instances where a driver’s score in the cooperative game framework surpasses their score in the non-cooperative framework by more than 5\%, the driver is deemed to exhibit greater cooperation during interactions. Conversely, if the non-cooperative game scores exceed the cooperative game scores by more than 5\%, the driver is considered more competitive. If neither score exceeds the other by this margin, the driver is regarded as demonstrating a balance of both competitive and cooperative behaviors.

Table \ref{tab:percentage cooperation and competition} shows that 31.48\% of drivers from Xianxia are predominantly competitive, compared to 17.58\% of drivers from Waymo. Furthermore, 25.93\% of Xianxia drivers show a stronger inclination towards cooperation, closely matched by 26.37\% of Waymo drivers. A higher proportion of Waymo drivers (56.04\%) display a balanced mix of competition and cooperation, compared to 42.59\% of Xianxia drivers. Overall, Xianxia drivers appear more competitive, while Waymo drivers tend to exhibit stronger cooperative behaviors.

\begin{table}[!htbp]
    \centering
    \caption{The percentages of competition and cooperation tendency fro two datasets}
    \begin{tabular}{c c c}
        \toprule
        Type & Waymo Drivers & Xianxia Drivers \\
        \midrule
        Stronger Competition & 17.58 \%  & 31.48 \% \\
        The Balance	& 56.04 \%	& 42.59\% \\
        Stronger Cooperation &	26.37 \% &	25.93\% \\
        \bottomrule
    \end{tabular}
    \label{tab:percentage cooperation and competition}
\end{table}

\subsubsection{Case Analysis}
Three cases from different drivers are chosen to analyze the processes of action-taking and interaction with the non-cooperative and cooperative game models.

\begin{figure}[!htbp]
    \centering
    \includegraphics[width=1\textwidth]{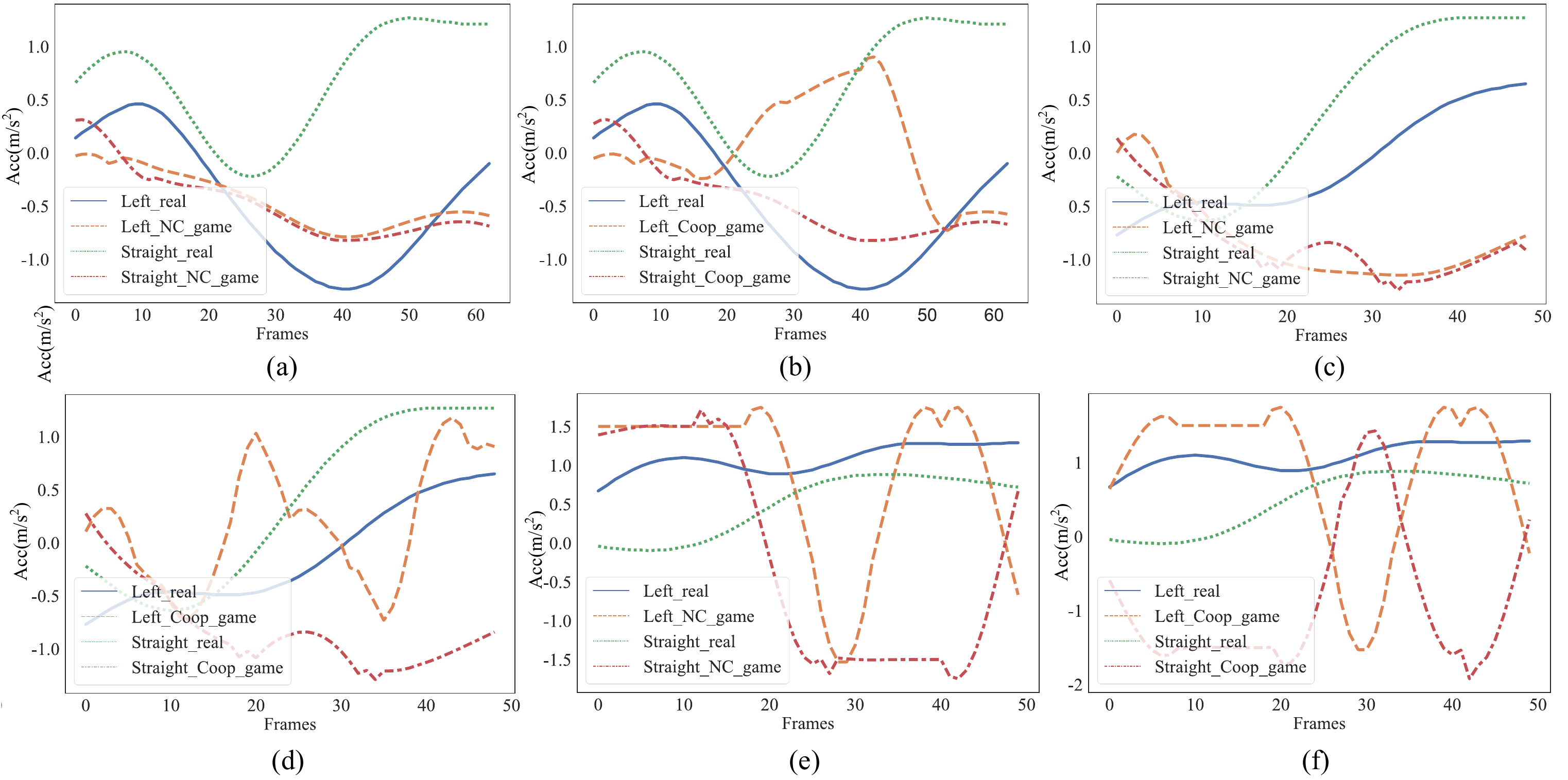}
    \caption{The results of acceleration from three cases with the non-cooperative game and cooperative game, (a),(b): Scenario 18; (c),(d) : Scenario 33; (e), (f) : Scenario 40.}
    \label{fig:Case_analysis}
\end{figure}

The acceleration information for the three cases is displayed in Figure \ref{fig:Case_analysis}. For each case, we discuss the real-world driver, the non-cooperative game results (in the left half of the subplot), and the cooperative game results (in the right half of the subplot).

In Case 1, as illustrated in Figure \ref{fig:Case_analysis} (a) and (b), the straight real-world driver traverses the intersection with minimal deceleration, while the left-turning real-world driver decelerates and stops to yield. However, both the left-turning and straight actions solved by the non-cooperative model involve deceleration, causing both drivers to wait for the other to pass first and leading to a dilemma. This may be due to both drivers in the non-cooperative game overestimating the safety risk of rushing and opting for conservative parking behavior. In contrast, the left-turning driver in the cooperative game agrees with the straight driver and begins to accelerate at the $20^{th}$ timestamp, after tentative actions spanning two seconds. This results in the cooperative game's solutions achieving higher global profit.

Similar features can be observed in Case 2, as shown in Figure \ref{fig:Case_analysis} (c) and (d). Both drivers in the non-cooperative game continuously decelerate, akin to Case 1. The straight driver and left-turning driver both initially decelerate and then accelerate, even though the straight driver's actions should take precedence. In the cooperative game, the left-turning driver alternates between acceleration and deceleration before ultimately speeding through the intersection. These more frequent action adjustments result in higher payoffs for both players but also lead to reduced comfort, as comfort is not considered in the game model proposed in this paper.

Case 3, depicted in Figure \ref{fig:Case_analysis} (e) and (f), presents a more pronounced and vivid negotiation and game process. Both real-world drivers exhibit minimal maneuver adjustments, while the drivers in the game model interact more frequently. When approaching the intersection, both drivers in the non-cooperative model switch from acceleration to deceleration after two seconds, after which the left-turning driver continues to accelerate. In the cooperative game, the actions of the two drivers are more dynamic. After the left-turning driver decelerates, the straight driver attempts to accelerate to adjust their motion states. Once the situation is clear, the left-turning driver begins to accelerate, and the straight driver decelerates accordingly.

Overall, drivers in the game model can react and modify their actions more quickly than real-world drivers. The cooperative game model enables drivers to achieve greater global benefits, highlighting the advantages and characteristics of cooperative games. Moreover, as seen in Figure \ref{fig:Case_analysis} (a) and (c), having two drivers who both exhibit stronger competition does not lead to better interactions, even if both drivers are completely rational.

\subsubsection{Discussion about of the practicality of our framework}
In this subsection, we examine the practicality and transferability of our framework. We have developed a universal evaluation framework that transcends the confines of specific application scenarios, enabling its use across a variety of interaction environments such as lane changing, ramp merging, and navigating urban roadways. We selected unsignalized intersections for primary application due to the complex challenges posed by unprotected left turns, which are particularly difficult to model.

Our framework is characterized by its flexibility and scalability, which significantly enhances its broad applicability. Within the game-theoretical dynamic assessment benchmark we established, we primarily consider factors such as safety and efficiency. Nevertheless, the parameters and factors within this benchmark are designed to be dynamically adjustable and replaceable. This adaptability allows our framework to accommodate the requirements of varying scenarios and diverse operational demands. For future enhancements, we could integrate additional factors like driver comfort and behavioral consistency, providing a more thorough evaluation of interactions. Furthermore, the flexibility to modify the weights of different evaluation criteria enables our framework to meet diverse assessment needs efficiently.

\subsection{Comparison with the baseline method}

To demonstrate the reliability of our model, we compare it with the baseline evaluation index PET. Since PET can only evaluate the safety of interaction results, we replace the risk model in the evaluation framework with the PET indicator to obtain a baseline model, ensuring the rationality of the comparative analysis. We then analyze the results of both models.

When employing the PET metric, player $i$'s payoff function is as follows:
\begin{equation}
    ub_i^k (\alpha_i)=m_i^k E_i(k)+n_i^k \big(PET_i(k)\big)
\end{equation}
\begin{equation}
    PET_i(k) = \lvert \big(
    \frac{dis_i(k)}{v_i(k+1)+\epsilon}
    - \frac{dis_o(k)}{v_o(k+1)+\epsilon}
    \big) \rvert
\end{equation}
where driver $o$ is the target of interaction for driver $i$, $dis(k)$ represents the distance between player $i$ and conflict point at $k^{th}$ interval, $v(k+1)$ represents player's velocity at $k+1^{th}$ interval, $\epsilon$ is a minimum to ensure existence of the function.

\begin{figure}
    \centering
    \includegraphics[width=1 \textwidth]{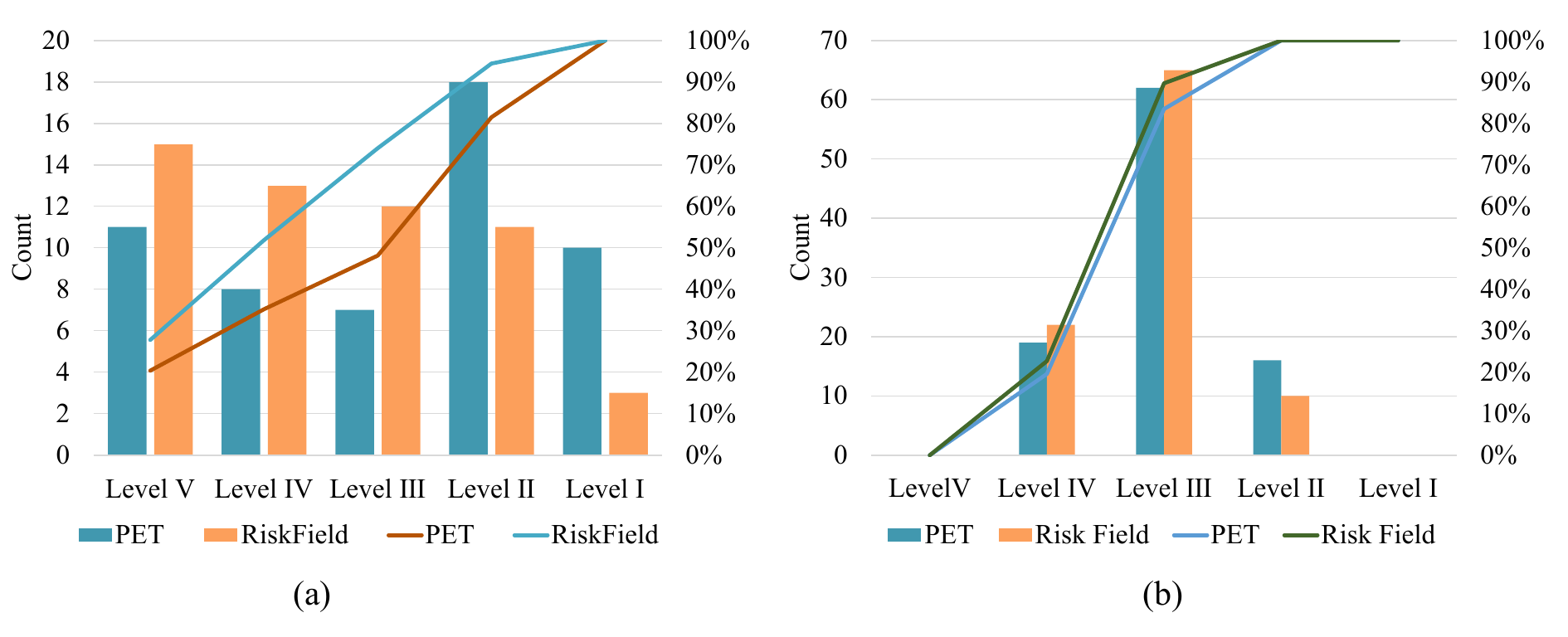}
    \caption{Comparison of the distribution of driving ability grades.(a) Xianxia-Jianhe Dataset; (b) Waymo Dataset.}
    \label{fig:PET_result}
\end{figure}

Figure \ref{fig:PET_result} presents the drivers' performance using two different safety payoff functions. Compared to the PET index, the safety ability scores calculated based on the risk field index are lower, indicating that the risk field index is more cautious and conservative than the PET index. In the Xianxia-Jianhe Dataset, the difference between scores from the two standards is substantial, whereas, in the Waymo Dataset, the difference is negligible, which may be attributed to the conservative behaviors of Waymo drivers.

\begin{figure}[!htbp]
    \centering
    \includegraphics[width=0.8 \textwidth]{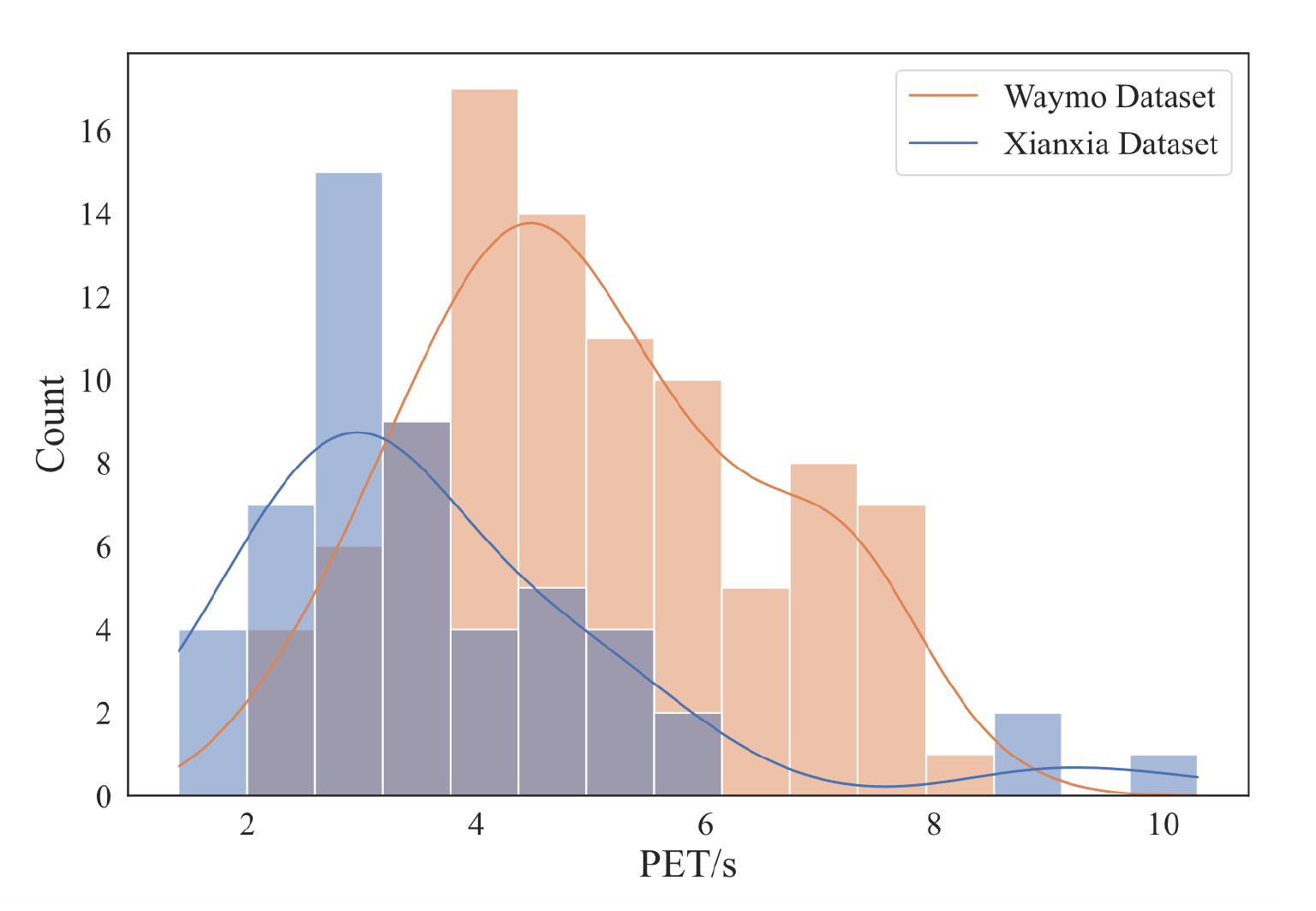}
    \caption{PET frequency distribution of interaction events from the two datasets.}
    \label{fig:PET_count}
\end{figure}

Furthermore, we calculated and analyzed the Post Encroachment Time (PET) for all interaction events in both datasets. Generally, smaller PET values indicate more aggressive interactions and higher security risks. The distribution and average PET values are illustrated in Figure \ref{fig:PET_count}. The average PET in the Xianxia Dataset is smaller than in the Waymo Dataset, suggesting more aggressive driving behaviors. We confirmed the statistical significance of the differences in data distribution between the two datasets using a t-test, as shown in Table \ref{tab:PET Descrition Statistics} and Table \ref{tab:Independent T-Test}. This analysis reveals that, even based solely on PET, drivers in the Xianxia dataset demonstrate higher aggression and efficiency but lower safety performance compared to Waymo drivers. These findings align with those derived from our interactive evaluation framework.

\begin{table}[!htbp]
    \centering
    \caption{Description Statistics of PET from two datasets.}
    \begin{tabular}{c c c c}
        \toprule
        Group  & Mean/s & Std. Deviation & Std. Error Mean \\
        \midrule
        Xianxia Drivers  & 3.6906 & 1.8081 & 0.2484 \\
        Waymo Drivers  & 4.9913 & 1.4950 &0.1559 \\
        \bottomrule
    \end{tabular}
    
    \label{tab:PET Descrition Statistics}
\end{table}

\begin{table*}[!htbp]
    \centering
    \caption{Independent T-Test results of two datasets.}
    \resizebox{\textwidth}{10mm}{
    \begin{tabular}{*{8}{c}}
        \toprule
        \multirow{2}*{ } & \multicolumn{2}{c}{Levene's Test for Equality of Variances} & \multicolumn{5}{c}{T-test for Equality of Means} \\
        \cmidrule(lr){2-3}\cmidrule(lr){4-8}
        {} & F & Sig. & t & df & Sig. (2-tailed) & Mean Difference & Std. Error Difference \\
        \midrule
        Equal variances assumed & 0.127 & 0.722 & -4.668 & 143 & 0 & -1.30074 & 0.27865 \\
        Equal variances not assumed & & &  -4.436 & 92.796 & 0 & -1.30074 & 0.29322 \\
        \bottomrule
    \end{tabular}
    }
    
    \label{tab:Independent T-Test}
\end{table*}

\section{Conclusions}
\label{section:conclu6}
In mixed human-machine driving environments, analyzing human drivers' interaction behaviors and assessing their interactive capabilities are crucial for enhancing the decision-making abilities of autonomous vehicles. Our work introduces a framework for assessing interaction capabilities oriented towards the interactive process, which includes three components: Risk Perception Modeling, Interactive Process Modeling, and Interactive Ability Scoring. This framework quantifies interaction risks through the estimation of interaction motion states and the application of risk field theory. The interaction behavior of drivers is modeled using various game-theoretical models, and the safety profits of these models are evaluated by the Risk Perception module. The action outputs from these game models represent rational agents' actions, serving as the benchmark for ability evaluation. We assess drivers' abilities by measuring the difference in actions between drivers and these rational agents. Additionally, we propose an improved evaluation index based on morphological similarity distance to score and assess drivers' real-world capabilities across three dimensions: safety, efficiency, and interactive sociality.

We applied our framework in scenarios involving unsignalized intersections and analyzed typical behavior datasets from drivers in China and the USA. The results indicate that our framework can effectively identify and assess drivers' interaction styles and tendencies. It effectively distinguishes between conservative and aggressive interaction behaviors and demonstrates good adaptability across different regional contexts.

\section{Disclosure statement}
No potential conflict of interest was reported by the author(s).

\section{Funding}
This work was supported in part by the National Natural Science Foundation of China (52232015), in part by the Fundamental Research Funds for the Central Universities (No. 2022-5-ZD-02 and No. 22120220434), and in part by the Zhejiang Laboratory (2021NL0AB02).

\bibliographystyle{tfcad}
\bibliography{reference}

\end{document}